\documentclass[10pt, conference, compsocconf]{IEEEtran}
\usepackage{cite}

\ifCLASSINFOpdf
  \usepackage[pdftex]{graphicx}
  \graphicspath{{../pdf/}{../jpeg/}}
  \DeclareGraphicsExtensions{.pdf,.jpeg,.png}
\else
  \usepackage{graphicx}
  \graphicspath{{../eps/}}
\fi
\usepackage[cmex10]{amsmath}
\usepackage{multirow}
\begin{document}
%
\title{Logical Parsing from Natural Language Based on a Neural Translation Model }




%
\author{\IEEEauthorblockN{Liang Li\IEEEauthorrefmark{1},
Pengyu Li\IEEEauthorrefmark{1},
Yifan Liu\IEEEauthorrefmark{1},
Tao Wan\IEEEauthorrefmark{2} and
Zengchang Qin\IEEEauthorrefmark{1}}
\IEEEauthorblockA{\IEEEauthorrefmark{1}Intelligent Computing and Machine Learning Lab, School of ASEE\\
\IEEEauthorblockA{\IEEEauthorrefmark{2}Biomedical Imaging and Informatics Lab, School of Biological Science and Medical Engineering\\}
Beihang University,
Beijing 100191, China\\ 
\{liangli919, lipengyu, yifan\_liu, taowan, zcqin\}@buaa.edu.cn
}

}


\maketitle

\begin{abstract}
Semantic parsing has emerged as a significant and powerful paradigm for natural language interface and question answering systems. Traditional methods of building a semantic parser rely on high-quality lexicons, hand-crafted grammars and linguistic features which are limited by applied domain or representation. In this paper, we propose a general approach to learn from denotations based on Seq2Seq model augmented with attention mechanism. We encode input sequence into vectors and use dynamic programming to infer candidate logical forms. We utilize the fact that similar utterances should have similar logical forms to help reduce the searching space. Under our learning policy, the Seq2Seq model can learn mappings gradually with noises. Curriculum learning is adopted to make the learning smoother. We test our method on the arithmetic domain which shows our model can successfully infer the correct logical forms and learn the word meanings, compositionality and operation orders simultaneously.
\end{abstract}


%
\IEEEpeerreviewmaketitle

\section{Introduction}
The problem of learning a semantic parser has been receiving significant attention. Semantic parsers map natural language into a logical form that can be executed on a knowledge base and return an answer (denotation). The early works use logical forms as supervision \cite{Miller2002A,Ge2005A,Lu2008A,Zettlemoyer2010Learning}, given a set of input sentences and their corresponding logical forms, learning a statistical semantic parser by weighting a set of rules mapping lexical items and syntactic patterns to their logical forms. Given an input, these rules are applied recursively to derive the most probable logical form. However, the tremendous labor needed for annotating logical forms has turned the trend to weak supervision $-$ using denotation of logical forms as the training target. It has been successfully applied in different fields including question-answering \cite{Clarke2010Driving,Liang2011Learning,Krishnamurthy2012Weakly,Reddy2014Large} and robot navigation \cite{Artzi2013Weakly}. All these works need hand-crafted grammars that are crucial in semantic parsing but pose an obstacle for generalization. Wang \emph{et al.} \cite{Wang2015Building} build semantic parsers in 7 different domains and hand engineer a separate grammar for each domain.

The rise of Seq2Seq model \cite{Sutskever2014Sequence} provides an alternative method to tackle the mapping problem and no more manual grammars are needed. The ability to deal with sequences with changeable length as input and/or  output has translated this model into applications including machine translation \cite{Sutskever2014Sequence, Cho2014Learning}, syntactic parsing \cite{Vinyals2014Grammar}, and question answering \cite{Hermann2015Teaching}. All of these work do not need hand-crafted grammars and are so-called end-to-end learning. But they do not resolve the problem of supervision by denotation, which makes one step further and needs logic reasoning and operation. Our model adopts the encoder-decoder framework and tries to use denotation as the target of supervised learning. We take the advantage of the Seq2Seq model's ability of tackling input/output with different length and grammar-free form to learn the mappings from natural language to logical forms. Our main focus is to infer logical forms from denotations in a generalizable way. We wish to add minimal extra constraints or manual features, so that it can be applied to other domains. For now, we can not infer correct logical forms all the time but we prove that the Seq2Seq model is capable of learning with noises in training data and the curriculum learning form mitigates this effect.

A problem of weak supervision is the search of the consistent logical forms when only denotation is available. The number of logical forms grows exponentially as their size increases and inferring from denotations inevitably induces spurious logical forms $-$ those that do not represent the original sentence semantics but get the correct answer accidently \cite{Pasupat2016Inferring}. To control the searching space, previous works relied on restricted sets of rules which limits expressivity and are possible to rule out the correct logical form. Instead, we utilize dynamic programming to construct the candidate logical form set with an extra base case set to filter all the candidates, and use the ongoing training Seq2Seq model to determine the best one. In this way, we can maintain the expressivity and train our model iteratively by feeding the suggested best logical form back to our model. Consequently, a right pick results in a positive learning, which allows our model to put a higher probability on the correct logical form over others and leads to the desired mapping over training.

We evaluate our model on arithmetic domain through a toy example. And the model is capable of translating row sentences into mathematical equations in structured tree form and returning the answer directly. We provide a base case set which serves as a pivot for the model to learn.Our model learns the arithmetic calculation in a curriculum way, where simpler sentences with fewer words are inputted at initial state. Our model assumes no prior linguistic knowledge and learns the meaning of all the words, compositionality, and order of operations from just the natural language $-$ denotation pairs.

\subsection{Related work}
We adopt the general encoder-decoder framework based on neural networks augmented with attention mechanism \cite{Bahdanau2014Neural}, which allows the model to learn soft alignment between utterances and logical forms. Our work is related to \cite{Dong2016Language} and \cite{Vinyals2014Grammar}, both of which use the Seq2Seq model to map natural language to logical forms in tree structure without hand-engineered features. But our work makes one step further by using denotation of logical form as the learning target and regard logical forms as latent variables.

How to reduce the searching space is a chief challenge in weak supervision. A common approach is to constrain the set of possible logical form compositions, which can significantly reduce the searching space but also constrain the expressivity \cite{pasupat2015compositional}.   Lao \emph{et al.} \cite{lao2011random} use random walks to generate logical forms and use denotation to cut down the searching space during learning. Liang \emph{et al.} \cite{Liang2016Neural} adopt a similar method to narrow down the options and allow more complex semantics to be composed. Different from generating logical forms forwardly as mentioned above, an alternative method is to use denotation to infer logical forms using dynamic programming \cite{Lau2003Programming,Pasupat2016Inferring}. In this way, it is more likely to recover the full set and find the desired one. Inspired by their work, we also employ this method and store the denotation $-$ logical form pairs in advance to accelerate the lookup efficiency.

Our work is similar to \cite{Liang2015Bringing} in the sense that we both focus on the arithmetic domain and learn from denotations directly. But their work relies on hand-crafted grammars to construct logical forms and hand engineered features to filter out incorrect logical forms. Instead, the Seq2Seq model we use is grammar-free and the features we select to screen logical forms is more general. Our main idea is that similar utterance should have similar logical forms, so that we can filter out incorrect logical forms by using similarity measurement. We build up a small base case set to assist this idea and the similarity function we adopt is simply bag-of-words, which is replaceable when extending to other fields. Furthermore, to sort the candidates, we do not have a particular scoring function to weight extracted features, instead, we use the ongoing training Seq2Seq model to evaluate their loss.

There are some other related work, Neural Programmer \cite{Neelakantan2015Neural} augmented with a small set of arithmetic and logic operations is able to perform complex reasoning and has shown success in question answering \cite{Neelakantan2016Learning}. Neural Turing Machines \cite{Graves2014Neural} can infer simple algorithms such as copying and sorting with external memory.

\begin{table*}[!tbp]
\centering
\caption{The base case set with several pairs of $<u,s>$, helping to reduce searching space}
\begin{tabular}{|c|c|c|c|c|c|}
\hline
 Utterance & Logical form & Denotation\\
\hline
one plus two $\langle$ eos $\rangle$ & $\langle$ `Go' `[' `1' `+' `2' ']' 'End' $\rangle$ & 3.0\\
three minus four times five $\langle$ eos $\rangle$ & $\langle$ `Go' `[' `3' `-' `(' `4' `*' `5' `)' `]' `End' $\rangle$ & -17.0\\
one times two divide three $\langle$ eos $\rangle$ & $\langle$ `Go' `(' `(' `1' `*' `2' `)' `/' `3' `)' `End' $\rangle$ & 0.667\\
two divide four plus five $\langle$ eos $\rangle$ & $\langle$ `Go' `[' `(' `2' `/' `4' `)' `+' `5' `]' `End' $\rangle$ & 5.5\\
five divide one times two plus three $\langle$ eos $\rangle$ & $\langle$ `Go' `[' `(' `(' `5' `/' `1' `)' `*' `2' `)' `+' `3' `]' `End' $\rangle$ & 13.0\\
four minus two times three plus one $\langle$ eos $\rangle$ & $\langle$ `Go' `[' `[' `4' `-' `(' `2' '*' `3' `)' `]' `+' `1' `]' `End' $\rangle$ & -1.0\\
three divide four minus five plus two $\langle$ eos $\rangle$ & $\langle$ `Go' `[' `[' `(' `3' `/' `4' `)' `-' `5' `]' `+' `2' `]' 'End' $\rangle$ & -2.25\\
\hline
\end{tabular}
\end{table*}
\section{Background:sequence-to-sequence model and attention mechanism}
Before introducing our model, we describe briefly the Seq2Seq model and attention mechanism.

\subsection{Sequence-to-sequence model}
The Seq2Seq model takes a source sequence $X=(x_1,x_2,...,x_T)$ as input and outputs a translated sequence $Y=(y_1,y_2,...,y_{T'})$. The model maximizes the generation probability of $Y$ conditioned on $X$: $p(y_1,...,y_{T'}|x_1,x_2,...,x_T)$. Specifically, the Seq2Seq is in an encoder-decoder structure. In this framework, an encoder reads the input sequence word by word into a vector $c$ through recurrent neural network (RNN).

\begin{equation}h_t=f(x_t,h_{t-1})\end{equation}
and
$$c=q({h_1,...,h_T}),$$
where $h_t$ is the hidden state at time $t$, $c$ is commonly taken directly from the last hidden state of encoder $q({h_1,...,h_T})=h_T$, and $f$ is a non-linear transformation which can be either a long-short term memory unit (LSTM) \cite{Hochreiter1997Long} or a gated recurrent unit (GRU) \cite{Cho2014On}. In this paper, LSTM is adopted and is parameterized as
\begin{subequations}
\begin{equation}
\left( {\begin{array}{*{20}{c}}
{{i_t}}\\
{{f_t}}\\
{{o_t}}\\
{{{\tilde c}_t}}
\end{array}} \right) = \left( {\begin{array}{*{20}{c}}
\sigma \\
\sigma \\
\sigma \\
{\tanh }
\end{array}} \right)T\left( {\begin{array}{*{20}{c}}
{{x_t}}\\
{{h_{t - 1}}}
\end{array}} \right)
\end{equation}

\begin{equation}
{c_t} = {f_t} \circ {c_{t - 1}} + {i_t} \circ {{\tilde c}_t}
\end{equation}
\begin{equation}
{h_t} = {o_t} \circ \tanh ({c_t})
\label{2c}
\end{equation}
\end{subequations}
where $\circ$ is an element-wise multiplication, $T$ is an affine transformation, $\sigma$ is the logistic sigmoid that restricts its input to [0,1], $i_t$, $f_t$ and $o_t$ are the input, forget, and output gates of the LSTM, and $c_t$ is the memory cell activation vector. The forget and input gates enable the LSTM to regulate the extent to which it forgets its previous memory and the input, while the output gate controls the degree to which the memory affects the hidden state. The encoder employs bidirectionality, encoding the sentences in both the forward and backward directions, an approach adopted in machine translation \cite{Bahdanau2014Neural,Cho2014Learning}. In this way, the hidden annotations ${h_t} = (\overrightarrow h _t^T;\overleftarrow h _t^T)$ concatenate forward $\overrightarrow h _t^T$ and backward annotations $\overleftarrow h _t^T$ together, each determined using Equation \ref{2c}.

The decoder is trained to estimate generation probability of next word $y_t$ given all the previous predicted words and the context vector $c$. The objective function of Seq2Seq can be written as \begin{equation} p(y_1,...,y_{T'}|x_1,...,x_T)=\prod_{t=1}^{T'}p(y_t|c,y_1,...,y_{t-1}).\label{2}\end{equation}With an RNN, each conditional probability is modeled as \begin{equation} p(y_t|{y_1,...,y_{t-1}},c)=g(y_{t-1},s_t,c),\label{3}\end{equation} where $g$ is a non-linear function that outputs the probability of $y_t$, $y_{t-1}$ is the predicted word at time $t-1$ in the response sequence, and $s_t$ is the hidden state of the decoder RNN at time $t$, which can be computed as
\begin{equation}s_t=f(y_{t-1},s_{t-1},c).\label{4}\end{equation}

\subsection{Attention mechanism}
The traditional Seq2Seq model predicts each word from the same context vector $c$, which deprives the source sequence information and makes the alignment imprecisely. To address this problem, attention mechanism is introduced to allow decoder focusing on the source sequence instead of a compressed vector upon prediction \cite{Bahdanau2014Neural}. In Seq2Seq with attention mechanism, each $y_i$ in $Y$ corresponds to a context vector $c_i$ instead of $c$. The conditional probability in Equation \ref{3} becomes
\begin{equation}p(y_i|y_1,...,y_{i-1},x)=g(y_{i-1},s_i,c_i),\label{5}\end{equation}
where the hidden state $s_i$ is computed by
\begin{equation}s_i=f(y_{i-1},s_{i-1},c_i).\end{equation}
The context vector $c_i$ is a weighted average of all hidden states $\{ {h_t}\} _{t = 1}^T$ of the encoder, defined as
\begin{equation}c_i=\sum_{j=1}^{T}\alpha_{ij}h_j,\end{equation}
where the weight $\alpha_{ij}$ is given by
\begin{equation}\alpha_{ij}=\frac{\exp(e_{ij})}{\sum_{k=1}^{T} \exp(e_{ik})},\end{equation}
where $e_{ij}$ is an alignment model which scores how well the inputs around position $j$ and the output at position $i$ match,
\begin{equation}e_{ij}=a(s_{i-1},h_j),\end{equation}
where $a$ is a feed forward neural network, trained jointly with other components of the system.

\begin{figure*}[!t]
\centering
\includegraphics[width=1\textwidth]{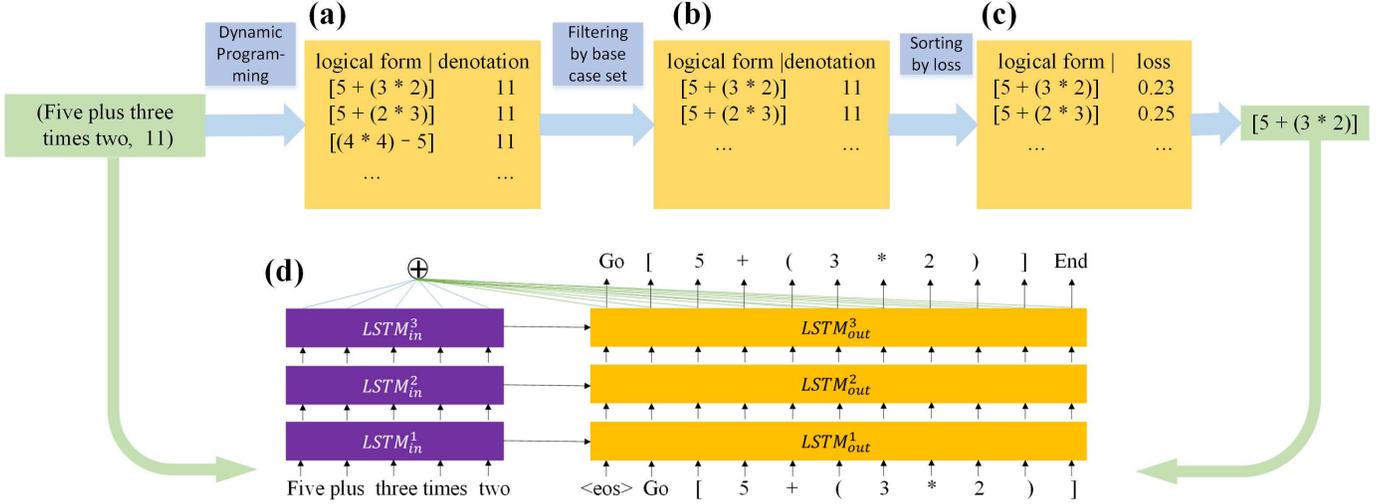}
\hfil
\caption{Learning algorithms applied to one example. The utterance is "Five plus three times two" with denotation 11. In (a), we show the inferred logical forms from the denotation using dynamic programming. In (b), the candidate logical forms are filtered by base case set. In (c), we use the ongoing Seq2Seq model to rank the loss of each candidate logical form and return the least one for training. (d) shows Seq2Seq model with attention mechanism with 3-layer LSTM on both encoder and decoder side.}
\end{figure*}

\section{Learning from denotations}
We use the triple $\langle u,s,d \rangle$ to denote the linguistic objects, where $u$ is an utterance, $s$ is a logical form and $d$ is the denotation of $s$. We use $\lfloor u\rfloor$ to represent the translation of utterance into its logical form, and we use $ [\kern-0.15em[ s ]\kern-0.15em] $ for the denotation of logical form $s$. Each training data is composed by the pair $\langle u,d \rangle$ without explicitly telling its correct logical form $s$.

With denotation as target label of learning, the Seq2Seq model is trained to put a high probability on $\lfloor u\rfloor$'s that are consistent-logical forms that execute to the correct denotation $d$. When the space of logical forms is large, searching for the correct logical form could be cumbersome. Additionally, different from the previous study which incorporates prior knowledge such as word embedding \cite{Liang2016Neural}, object categories \cite{Artzi2013Weakly}, our model in this mathematical expression learning example has no such knowledge and has to learn the meanings of input utterance.

Here we formally describe the methods to reduce the searching space and how to infer the correct logical forms from denotations.

\subsection{Dynamic programming on denotations}
Our first step is to generate all logical forms that have the correct denotations. Formally, given a denotation $d$, we wish to generate a candidate logical form set that satisfy the denotation demand $\Omega  = \{ s|[\kern-0.15em[ s ]\kern-0.15em]=d\}.$ Previous work use beam search to generate candidates but it is hard to recover the full set $\Omega$ due to pruning. Noticing that one denotation may correspond to multiple logical forms, which leads to the increase of the number of distinct denotations is much slower than the number of logical forms. We use dynamic programming on denotations to recover the full set, following the work of \cite{Pasupat2016Inferring,Lau2003Programming}. A necessary condition for dynamic programming to work is denotationally invariant semantic function $g$, such that the denotation of the resulting logical form $g(s_1,s_2)$ only depends on the denotations of $s_1$ and $s_2$. In the arithmetic domain, the result of an equation can be computed recursively and independently which certainly satisfies this requirement.

Our primary purpose is to collapse logical forms with the same denotation together, so that given a denotation $d$ the candidate set $\Omega$ (Figure 1. (a)) can be returned directly. In order to speed up the lookup efficiency, we store the pair $(d,\Omega)$ in advance.

\subsection{Filter candidate logical forms}
In order to filter out incorrect logical forms, we utilize the fact that similar utterances should have similar logical forms to reduce the searching space. We build up a base case set $B$ (Table I) that stores several $\langle u_b,s_b \rangle$ pairs with varying utterance length. Specifically, given an input pair $\langle{u_i,d_i}\rangle$, we iterate the base case set to find the one which shares the most similarity with input utterance $u_i$.

\begin{equation}{\tilde u_b} = \mathop {\arg \min }\limits_{{u_b} \in B} e(\phi ({u_b}),\phi ({u_i})),\end{equation}
where $\phi$ extracts features from utterance $u_b$ and $u_i$, here we use bag-of-words as feature extraction function, and we simply counts the shared features as the feature similarity measurement function $e$.

After finding the base case utterance ${\tilde u_{b}}$ which is closest to the input utterance $u_i$, we use the corresponding base case logical form ${\tilde s_{b}}$ to filter the candidate set $\Omega$.

\begin{equation}
\Gamma  = \{ s|s = \mathop {\arg \max }\limits_{{s_j} \in \Omega } \frac{{e(\phi ({{\tilde u}_b}),\phi ({u_i}))}}{{e(\phi ({{\tilde s}_b}),\phi ({s_j}))}}\}.
\end{equation}

We use a new set $\Gamma$ (Figure 1. (b)) to store these updated candidate logical forms that have similar features with the base case $\langle\tilde u_{b},\tilde s_{b}\rangle$. To further determine the most probable one from set $\Gamma$, we use the Seq2Seq model to examine the loss of each candidate and select the one with least loss to return for training.
\begin{equation}{\tilde s_i} = \mathop {\arg \max }\limits_{s \in \Gamma } {p_{Seq2Seq}}(s)\end{equation}

The selected logical form $\tilde s_i$ is paired with its utterance to form a training example $\langle{u_i,\tilde s_i}\rangle$ which feeds back to the Seq2Seq model for training (Figure 1. (c) - (d)).

\section{Experimental studies}

\subsection{Dataset}
For the experiments, we randomly generate 8000 utterances with varying length from 3 to 7, split into a training set of 6000 training examples and 2000 test examples. Each utterance consists of integers from one to five and four operators `plus', `minus', `times', `divide'. Every utterance represents a legal arithmetic expression. Even the scope is quite limited, the searching space for an equation with length equal to seven is still numerous: $5^4*4^3=4*10^4$, which is caused by the rich compositionality of arithmetic equations. Besides, this nature gives rise to one denotation can correspond to much more logical forms compared with other domains, resulting in increasing noises for inference.

We select two ways to represent logical forms, both of them are represented by Arabic numerals and executable operators, but one is inserted with brackets to denote the calculation order, linearized from tree structure, the other assumes knowing calculation order without brackets for denotation. The base case set consists of seven samples, the brackets are omitted directly when considering logical forms with calculation knowledge (Table I).

The involvement of brackets largely increases the meaning of our logical form, for the reason that it actually represents tree structure with logic reasoning. Besides, not only has the model to learn soft alignment for brackets which are not introduced in the utterance explicitly, but it has to learn the brackets matching relationship.

\subsection{Settings}

For the convenience of data preprocessing and vectorization, each input sentence is appended an ending mark $\langle{eos}\rangle$, noting the end of input. Also, we manually set a maximum length for the input and output sentence, where the blank position will be automatically filled by `PAD' mark.

We construct 3 layers of LSTM on both encoder and decoder side with 20 hidden units for each layer. An embedding and a softmax layer is inserted as the first and last layer. Dropout is used for regularizing the model with a constant rate 0.3. Dropout operators are used between different LSTM layers and for the hidden layers before the softmax classifier. This technique can significantly reduce overfitting, especially on datasets of small size. Dimensions of hidden vector and word embedding are set to 20.

We use the RMSProp algorithm to update parameters with learning rate 0.001 and smoothing constant 0.9. Parameters are randomly initialized from a uniform distribution $\mathcal{U}(-0.05,0.05)$. We run the model for 200 epochs.

\subsection{Results and analysis}
We report the results with the Seq2Seq model on two variants, i.e., with brackets noting calculation order and without brackets as logical forms. To compare the performance of weak supervision, we also test the performance of traditional training with gold standard logical form. The result is reported on the accuracy of denotation $-$ the portion of input sentences are translated into the correct denotation. Table II presents the comparison.

\begin{table}[!h]
\caption{Overall accuracy on denotation}
\begin{tabular}{|c|c|c|}
\hline
$\multirow{2}{*}{Method}$                   & \multicolumn{2}{|c|}{Seq2Seq} \\
\cline{2-3}
                                                                & without brackets   & with brackets\\ \hline
Train with logical form     & 100.0\%             & 92.4\%\\
Train with denotation\hspace{\fill}          & 72.7\%             & 69.2\%\\ \hline
\end{tabular}
\end{table}

Overall, the accuracy with logical forms as supervision is much higher than with denotations, for the reason that training by gold logical forms does not bring any noises. On the other hand, the spurious logical forms affect our model's performance unavoidably. For instance, an utterance ``Five plus three times four" can be mistakenly translated into $[5+(4*3)]$, which has the correct denotation. This is reasonable because we adopt bag-of-words to measure similarity and this method does not take the order into consideration. Besides, not all of the training logical forms returned by our inference process are correct, but at least they have the correct denotation. So it is noticeable that even only approximately 45\% logical forms (Figure 2) returned for training are correct, the accuracy on denotation of our model is much higher than this limit. 

To find the best logical form for training, the final pick decision is made by the ongoing training Seq2Seq model, this training-by-prediction policy makes the model difficult to converge (Figure 2). But we can see this policy is effective to correct its previous prediction, which we call the ability of self-correction. This proves Seq2Seq model can learn word meanings and compositionality with noises.
\begin{figure}[!h]
\centering
\includegraphics[width=0.5\textwidth]{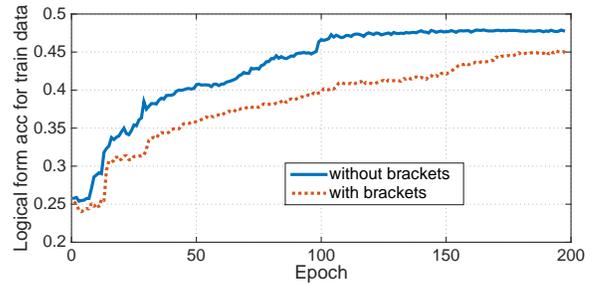}
\hfil
\caption{The percentage of correct logical form returned for training as the number of epoch increases.}
\end{figure}

In addition, the performance of training on logical forms with brackets is inferior to the model without brackets, which makes sense that longer sequence add difficulty for learning and brackets introduce more complicated mapping relationships. However, by utilizing soft alignment, the model can learn tree structure successfully with only denotation as supervision.

\section{Conclusion}
In this paper, we present an encoder-decoder neural network model for mapping natural language to their meaning representations with only denotation as supervision. We use dynamic programming to infer logical forms from denotations, and utilize similarity measurement to reduce the searching space. Also, curriculum learning strategy is adopted to smooth and accelerate the learning process. Under the policy training-by-predictions, our model has the ability of self-correction. We apply our model to the arithmetic domain and experimental results show that this model can learn word meanings and compositionality without resources to domain- or representation-specific features. One major problem remained in our work is that the model may confuse the order of predictions, which is caused by the inherent weakness of bag-of-words similarity measurement. This could be enhanced by some sequence-based similarity measurement in future work.

Although the example we test is rather simple, the expansibility to other fields and application scenarios is promising due to the few hand-engineered features and its capability of learning structured form. It would be interesting to learn a question-answering model with only question-answer pairs, or apply it to robot navigation task. We expect to extend our model to these fields and continue to enrich it.
\\





\bibliographystyle{IEEEtran}
%



\bibliography{bibfile}

\end{document}